# A Definition and Graphical Representation for Causality

David Heckerman
Microsoft Research, Bldg 9S/1
Redmond WA 98052-6399
heckerma@microsoft.com

Ross Shachter
Department of Engineering-Economic Systems
Stanford, CA 94305-4025
shachter@camis.stanford.edu

## Abstract

We present a precise definition of cause and effect in terms of a fundamental notion called unresponsiveness. Our definition is based on Savage's (1954) formulation of decision theory and departs from the traditional view of causation in that our causal assertions are made relative to a set of decisions. An important consequence of this departure is that we can reason about cause locally, not requiring a causal explanation for every dependency. Such local reasoning can be beneficial because it may not be necessary to determine whether a particular dependency is causal to make a decision. Also in this paper, we examine the graphical encoding of causal relationships. We show that influence diagrams in canonical form are an accurate and efficient representation of causal relationships. In addition, we establish a correspondence between canonical form and Pearl's causal theory.

## 1 Introduction

Most traditional models of uncertainty, including Markov networks (Lauritzen, 1982) and Bayesian networks (Pearl, 1988) have focused on the associational relationship among variables as captured by conditional independence and dependence. Associational knowledge, however, is not sufficient when we want to make decisions under uncertainty. For example, although we know that smoking and lung cancer are probabilistically dependent, we cannot conclude from this knowledge that we will decrease our chances of getting lung cancer if we stop smoking. In general, to make rational decisions, we need to be able to predict the effects of our actions.

Recent work by Artificial Intelligence researchers, statisticians, and philosophers—for example, Pearl and Verma (1991), Druzdzel and Simon (1993), and Spirtes et al. (1993)—have emphasized the importance of identifying causal relationships for purposes of modeling the effects of intervention. They argue, for example, that if we believe that smoking causes lung cancer, then we believe that our choice whether to continue or quit smoking can affect whether we get lung cancer. In contrast, if we believe that smoking does not cause lung cancer, our choice will not affect whether we get lung cancer, and the observed correlation between smoking and lung cancer could be explained perhaps by a common cause of both (e.g., a genetic predisposition toward cancer and the desire to smoke), which we are unable to control.

This recent work has led to significant breakthroughs in causal reasoning. For example, Pearl and Verma (1991) and Spirtes et al. (1993) have shown how causal knowledge represented graphically can be used to predict the effects of interventions and how observational data can be used to suggest causal relationships, and Pearl (1995) has shown how, given a qualitative causal structure, the quantitative effects of intervention may be estimated from observational data alone in some situations.

In this paper, we offer three improvements to the current work in causal reasoning. First, the current approaches either take causality as a primitive notion, or provide only a fuzzy, intuitive definition of cause and effect. For example, in the introduction of their book on causation, Spirtes et al. (1993, p. 42) write:

> We understand causation to be a relation between particular events: something happens and causes something else to happen. Each cause is a particular event and each effect is a particular event. An event $A$ can have more than one cause, none of which alone suffice to produce $A$. An event $A$ can also be overde-termined: it can have more than one set of causes that suffice for $A$ to occur. We assume that causation is transitive, irreflexive, and antisymmetric.

In this paper, we offer a definition of causation in terms of a more fundamental relation that we call unresponsiveness. Our definition is precise, and can be used as an assessment aid when someone is having trouble determining whether or not a relationship is causal. Also, our definition can help people accurately communicate their beliefs about causal relationships. In addition, the definition facilitates the development of techniques for learning causal relationships from data (Heckerman, this proceedings).

Second, the current approaches require all relationships to be causal. That is, for any two probabilistically dependent events or variables $x$ and $y$ in a given domain, these methods require a user to assert either that $x$ causes $y$, $y$ causes $x$, or they are linked by a chain of causal relationships, such as when $x$ and $y$ share a common cause, or $x$ and $y$ are common causes of an observed variable. For example, Pearl and Verma's (1991) causal model is a directed acyclic graph (DAG), wherein every node corresponds to a variable and every arc from nodes $x$ to $y$ corresponds to the assertion that $x$ is a direct cause of $y$. When using a causal model to represent a domain, a causal explanations must hold for every dependency in the domain. Our definition of causation is local in that it does not require all relationships to be causal. This property can be advantageous when making decisions. Namely, given a particular decision problem, there may be no need to assign a causal explanation to all dependencies in the domain in order to determine a rational course of action. Consequently, our definition may enable a decision maker to reason more efficiently.

Third, we describe a special condition on an influence diagram known as canonical form and show how it can be used to represent causal relationships more efficiently than existing representations.

Our approach is consistent with several current methods for reasoning about causality, including Pearl's causal theory (Pearl and Verma, 1991; Pearl, 1995) and causal networks of Spirtes et al. (1993). In addition, our approach is consistent with the philosophy of decision analysis as described by Savage (1954) and refined by Howard (1990). Thus, our discussions here offer a means by which the two disciplines may begin to communicate and contribute to each other's work.

This paper is a sequel to that presented at last year's conference (Heckerman and Shachter, 1994). Here, we clarify and generalize many of the concepts in the previous paper, including those of unresponsiveness (formerly discussed in terms of fixed sets), mapping variable, cause, set decision, and canonical form.

## 2 Unresponsiveness

In this section, we introduce the notions of unresponsiveness and limited unresponsiveness, fundamental relations underlying causation.

Important to our discussion are several distinctions from classical decision theory as described by Savage (1954). In particular, we distinguish between *alternatives* (what Savage called "acts"), *realizations* (what Savage called "consequences"), and *possible states of the world*.[1] Savage describes and illustrates these concepts as follows:

> To say that a decision is to be made is to say that one or more [alternatives] is to be chosen, or decided on. In deciding on an [alternative], account must be taken of the possible states of the world, and also of the [realizations] implicit in each [alternative] for each possible state of the world. A [*realization*] is anything that may happen to the person.
>
> Consider an example. Your wife has just broken five good eggs into a bowl when you come in and volunteer to finish making the omelet. A sixth egg, which for some reason must either be used for the omelet or wasted altogether, lies unbroken beside the bowl. You must decide what to do with this unbroken egg. Perhaps it is not too great an oversimplification to say that you must decide among three [alternatives] only, namely, to break it into the bowl containing the other five, to break it into a saucer for inspection, or to throw it away without inspection. Depending on the state of the egg, each of these three [alternatives] will have some [realization] of concern to you, say that indicated by Table 1.

For purposes of our discussion, there are two points to emphasize from Savage's exposition. First, it is important to distinguish between that which we can control directly—namely, alternatives—and that which we can control only indirectly through choosing an alternative—namely, realizations. Second, once we choose an alternative, the realization that occurs is logically determined by the state of the world. Of course, this realization can be (and usually is) uncertain, because the state of the world is uncertain.

---

[1] We use the term "alternative" in place of "act", because the former is more commonly used today. We use the term "realization" in place of "consequence" because it avoids the connotation that we should necessarily care about a realization. That is, we often want to model realizations, even though we don't directly care about them. In using different terms for these concepts, however, we do not intend to change their meanings.

Table 1: An example illustrating alternatives, possible states of the world, and realizations. (Taken from Savage [1954].)

| state of the world | alternative | | |
|---|---|---|---|
| | break into bowl | break into saucer | throw away |
| good | six-egg omelet | six-egg omelet and a saucer to wash | five-egg omelet and one good egg destroyed |
| bad | no omelet and five good eggs destroyed | five-egg omelet and a saucer to wash | five-egg omelet |

Table 2: The four possible states of the world for a decision to continue or quit smoking.

| state of the world | alternative | |
|---|---|---|
| | continue | quit |
| 1 | cancer | no cancer |
| 2 | no cancer | no cancer |
| 3 | cancer | cancer |
| 4 | no cancer | cancer |

In the omelet story, the possible states of the world readily come to mind given the description of the problem. Furthermore, we can observe the state the world (i.e., the condition of the egg). In many if not most situations, however, the state of the world is unobservable; and we can only bring the possible states to mind by thinking about the alternatives and realizations. For example, suppose we have a decision to continue smoking or quit, and we model the realizations of getting cancer or not. These alternatives and realizations bring to mind four possible states of the world, as shown in Table 2. These possible states have no familiar names; and we simply label them with numbers. The actual state of the world is not observable, because, if we decide to quit, we won't know for sure what would have happened had we continued, and vice-versa. Nonetheless, given the alternatives and realizations in this problem, these states of the world are well defined.[2] Also, note that, as illustrated by this example, there can be more possible states of the world than either realizations or alternatives. In general, if we have a decision problem with $r$ realizations and $a$ alternatives, then we can distinguish as many as $r^a$ possible states of the world.

In practice, it is often cumbersome if not impossible to reason about a monolithic set of alternatives, possible states of the world, or realizations. Consequently, we typically describe each of these items in terms of a set of variables. We call the variables describing a set of realizations *chance variables*. For example, in the omelet story, we can describe the realizations in terms of three variables: (1) *number of eggs in the omelet?*[3] ($o$), having instances "zero," "five," and "six," (2) *number of good eggs destroyed?* ($g$), having instances "zero," "one," and "five," and (3) *saucer to wash?* ($s$), having instances "no" and "yes." That is, every realization corresponds to an assignment of an instance to each chance variable.

We call the variables describing a set of alternatives *decision variables* (or *decisions*, for short). For example, suppose we have a set of alternatives about how we are going to dress for work. In this case, we can describe our alternatives in terms of the decision variables (say) *shirt* ("plain" or "striped"), *pants* ("jeans" or "corduroy"), and *shoes* ("tennis shoes" or "loafers"). In this example and in general, every alternative corresponds to an assignment of an instance to each decision variable.

The description of possible states of the world in terms of component variables is a bit more complicated, and is not needed for our explication of unresponsiveness and limited unresponsiveness. We defer discussion of this issue to Section 4.1.

As a matter of notation, we use $D$ to denote the set of decisions that describe the alternatives for a decision problem, and lower-case letters (e.g., $d, e, f$) to denote individual decisions in the set $D$. Also, we use $U$ to denote the set of chance variables that describe the realizations, and lower-case letters (e.g., $x, y, z$) to denote individual chance variables in $U$. In addition, we use the variable $S$ to denote the state of the world (the instances of $S$ correspond to the possible states of the world).[4] Thus, any given decision problem—or *domain*, as we sometimes call it—is described by the variables $U$, $D$, and $S$.[5]

With this introduction, we can discuss the concept of limited unresponsiveness. To illustrate this concept, consider the following decision problem adapted from Rubin (1978). Suppose we are a physician who has to decide whether or not to recommend a treatment to a patient. Given our recommendation, the patient may or may not actually accept the treatment, and may or may not be cured as a result. Here, we use a single decision variable *recommendation* ($r$) to represent our alternatives (i.e., $D = \{r\}$), and two chance variables *taken?* ($t$) and *cured?* ($c$) to represent whether or not the patient actually accepts the treatment and whether or not the patient is cured, respectively (i.e., $U = \{t, c\}$).

---

[2]Howard (1990) discusses in detail what it means for possible states of the world to be well defined.

[3]To emphasize the distinction between chance and decision variables, we put a question mark at the end of the names of chance variables.

[4]We use an uppercase "$S$" to denote this single variable, because later we decompose $S$ into a set of variables.

[5]Sometimes, for simplicity, we leave $S$ implicit in the specification of a decision problem.

The possible states of the world for this problem are shown in Table 3. For example, consider the first row in the table. Here, the patient will accept the treatment if and only if we recommend it, and will be cured if and only if he takes the treatment. We describe this state by saying that the patient is a "complier" and is "helped" by the treatment. We discuss the description of these states in more detail in Section 4.1.

As indicated in the table, we have asserted that the last four states of the world are impossible (i.e., have a probability of zero). These last four states share the property that $t$ takes on the same instance for both alternatives, whereas $c$ does not. Thus, this decision problem satisfies the following property: in all of the states of the world that are possible, if $t$ is the same for the two alternatives, then $c$ is also the same. We say that $c$ is unresponsive to $r$ in states limited by $t$.

In general, suppose we have a decision problem described by variables $U$, $D$, and $S$. Let $X$ be a subset of $U$, and $Y$ be a subset of $U \cup D$. We say that $X$ is unresponsive to $D$ in states limited by $Y$ if we believe that, for all possible states of the world, if $Y$ assumes the same instance for any two alternatives then $X$ must also assume the same instance for those alternatives.

To be more formal, let $X[\mathbf{S}, \mathbf{D}]$ be the instance that $X$ assumes (with certainty) given the state of the world $\mathbf{S}$ and the alternative $\mathbf{D}$. For example, in the omelet story, if $\mathbf{S}$ is the state of the world where the egg is good, and $\mathbf{D}$ is the alternative "throw away," then $o[\mathbf{S}, \mathbf{D}]$ (the number of eggs in the omelet) assumes the instance "five." Then, we have the following definition.

**Definition 1 (Limited (Un)responsiveness)**
*Given a decision problem described by chance variables $U$, decision variables $D$, and state of the world $S$, and variable sets $X \subseteq U$ and $Y \subseteq D \cup U$, $X$ is said to be unresponsive to $D$ in states limited by $Y$, denoted $X \not\leftarrow_Y D$, if we believe that $\forall \mathbf{S} \in S, \mathbf{D_1} \in D, \mathbf{D_2} \in D$,*

$$Y[\mathbf{S}, \mathbf{D_1}] = Y[\mathbf{S}, \mathbf{D_2}] \Longrightarrow X[\mathbf{S}, \mathbf{D_1}] = X[\mathbf{S}, \mathbf{D_2}]$$

*$X$ is said to be responsive to $D$ in states limited by $Y$, denoted $X \leftarrow_Y D$, if it is not the case that $X$ is unresponsive to $D$ in states limited by $Y$. That is, if we believe that $\exists \mathbf{S} \in S, \mathbf{D_1} \in D, \mathbf{D_2} \in D$ such that*

$$Y[\mathbf{S}, \mathbf{D_1}] = Y[\mathbf{S}, \mathbf{D_2}] \text{ and } X[\mathbf{S}, \mathbf{D_1}] \neq X[\mathbf{S}, \mathbf{D_2}]$$

When $X$ is (un)responsive to $D$ in states limited by $Y = \emptyset$, we simply say that *$X$ is (un)responsive to $D$*. The notion of unresponsiveness is significantly simpler than that of limited unresponsiveness. In particular, when $Y = \emptyset$, the equalities on the left-hand-side of the implications in Definition 1 are trivially satisfied. Thus, $X$ is unresponsive to $D$ if we believe that, for all possible states of the world and all alternatives, $X$ assumes the same instance; and $X$ is responsive to $D$, if there is some possible state of the world where $X$ differs for two different alternatives.

As examples of responsive variables, consider the omelet story. Let $\mathbf{S}$ denote the state where the egg is good, and $\mathbf{D_1}$ and $\mathbf{D_2}$ denote the alternatives "break into bowl" and "throw away," respectively. Then, for the variable $o$ (*number of eggs in omelet?*), we have $o[\mathbf{S}, \mathbf{D_1}] =$ "six" and $o[\mathbf{S}, \mathbf{D_2}] =$ "five". Consequently, $o$ is responsive to $D$. In a similar manner, we can conclude that $g$ (*number of good eggs destroyed?*), and $s$ (*saucer to wash?*) are each responsive to $D$ as well.

Note that, if an chance variable $x$ is responsive to $D$, then—to some degree—it is under the control of the decision maker. Consequently, the decision maker can not observe $x$ prior to choosing an alternative for $D$. For example, in the omelet story, we can not observe any of the responsive variables $o$, $g$, or $s$ before choosing an alternative.

As an example of an unresponsive variable, suppose we add $S$ (the state of the world) as a variable to $U$. (E.g., in the omelet story, we can take $U$ to be $\{S, o, g, s\}$.) By Savage's definition of $S$, it must be unresponsive to $D$. Note that adding $S$ to $U$ creates no new states of the world.

The notions of unresponsiveness and limited unresponsiveness are closely related to concepts in counterfactual reasoning (e.g., as described by Lewis (1979)). In particular, when we determine whether or not a set of chance variables $X$ is unresponsive to decisions $D$, we essentially answer the query "Will the outcome of $X$ be the same no matter how we choose $D$?" Furthermore, when we determine whether or not $X$ is unresponsive to $D$ in states limited by $Y$, we answer the query "Will the outcome of $X$ be the same no matter how we choose $D$, if $Y$ will not change as a result of our choice?" Queries of this form are of examples counterfactual queries. One of the fundamental assumptions of our work presented here is that these queries are easily answered. In our experience, we have found that decision makers are indeed comfortable answering such restricted counterfactual queries.

The concepts of responsiveness and probabilistic independence are related, as illustrated by the following theorem.

**Theorem 1** *If a set of chance variables $X$ is unresponsive to a set of decision variables $D$, then $X$ is*

Table 3: A decision about recommending a medical treatment.

| S (state of the world) | r (recommendation) | | | |
|---|---|---|---|---|
| | take | | don't take | |
| | t (taken?) | c (cured?) | t (taken?) | c (cured?) |
| 1: complier, helped | yes | yes | no | no |
| 2: complier, hurt | yes | no | no | yes |
| 3: complier, always cured | yes | yes | no | yes |
| 4: complier, never cured | yes | no | no | no |
| 5: defier, helped | no | no | yes | yes |
| 6: defier, hurt | no | yes | yes | no |
| 7: defier, always cured | no | yes | yes | yes |
| 8: defier, never cured | no | no | yes | no |
| 9: always taker, cured | yes | yes | yes | yes |
| 10: always taker, not cured | yes | no | yes | no |
| 11: never taker, not cured | no | no | no | no |
| 12: never taker, cured | no | yes | no | yes |
| 13: (impossible) | yes | yes | yes | no |
| 14: (impossible) | yes | no | yes | yes |
| 15: (impossible) | no | no | no | yes |
| 16: (impossible) | no | yes | no | no |

*probabilistically independent of* $D$.

**Proof:** By definition of unresponsiveness, $X$ assumes the same instance for all alternatives in any possible state of the world. Consequently, we can learn about $X$ by observing $S$, but not by observing $D$. □

Nonetheless, the two concepts are not identical. In particular, the converse of Theorem 1 does not hold. For example, let us consider the simple decision of whether to bet heads or tails on the outcome of a coin flip. Assume that the coin is fair (i.e., the probabilities of heads and tails are both 1/2) and that the person who flips the coin does not know our bet. Here, the possible outcomes of the coin toss correspond to the possible states of the world. Further, let decision variable $b$ denote our bet, and chance variable $w$ describe the possible realizations that we win or not. In this situation, $w$ is responsive to $b$, because for both possible states of the world, $w$ will be different for the different bets. Nonetheless, the probability of $w$ is 1/2, whether we bet heads or tails. That is, $w$ and $b$ are probabilistically independent.

Limited unresponsiveness and conditional independence are less closely related than are their unqualified counterparts. Namely, limited unresponsiveness does not imply conditional independence. For example, in the medical-treatment story, $c$ (*cured?*) is unresponsive to $r$ (*recommendation*) in states limited by $t$ (*taken?*), but it is reasonable for us to believe that $c$ and $r$ are not independent given $t$, perhaps because there is some gene that—partially or completely—determines how a person reacts to both recommendations and treatment.

We can derive several interesting properties of limited unresponsiveness from its definition.

1. $X \not\leftarrow_Y D \iff \forall x \in X, x \not\leftarrow_Y D$
2. $X \not\leftarrow_W D \iff X \cup W \not\leftarrow_W D$
3. $X \not\leftarrow_D D$
4. $X \not\leftarrow_Y D \implies X \not\leftarrow_{Y \cup Z} D$
5. $X \not\leftarrow_{Y \cup Z} D$ and $Y \not\leftarrow_Z D \implies X \not\leftarrow_Z D$
6. $X \leftarrow_Z D$ and $W \not\leftarrow_Z D \implies X \leftarrow_{W \cup Z} D$

where $D$ is the set of decision variables in the domain, $X$ and $W$ are arbitrary sets of chance variables in $U$, and $Y$ and $Z$ are arbitrary sets of variables in $U \cup D$.

The proofs of these properties are straightforward. For example, consider property 5. Given $X \not\leftarrow_{Y \cup Z} D$, we have $\forall \mathbf{S} \in S, \mathbf{D_1} \in D, \mathbf{D_2} \in D$,

$$Y[\mathbf{S}, \mathbf{D_1}] = Y[\mathbf{S}, \mathbf{D_2}] \text{ and } Z[\mathbf{S}, \mathbf{D_1}] = Z[\mathbf{S}, \mathbf{D_2}]$$
$$\implies X[\mathbf{S}, \mathbf{D_1}] = X[\mathbf{S}, \mathbf{D_2}]$$

Given $Y \not\leftarrow_Z D$, we have $\forall \mathbf{S} \in S, \mathbf{D_1} \in D, \mathbf{D_2} \in D$,

$$Z[\mathbf{S}, \mathbf{D_1}] = Z[\mathbf{S}, \mathbf{D_2}] \implies Y[\mathbf{S}, \mathbf{D_1}] = Y[\mathbf{S}, \mathbf{D_2}]$$

Consequently, we obtain $\forall \mathbf{S} \in S, \mathbf{D_1} \in D, \mathbf{D_2} \in D$,

$$Z[\mathbf{S}, \mathbf{D_1}] = Z[\mathbf{S}, \mathbf{D_2}] \implies X[\mathbf{S}, \mathbf{D_1}] = X[\mathbf{S}, \mathbf{D_2}]$$

That is, $X \not\leftarrow_Z D$.

Other properties follow from these. For example, it is true trivially that $\emptyset \not\leftarrow_Y D$. Consequently, by Property 2, we know that $Y \not\leftarrow_Y D$. As another example, a special case of Property 4 is that whenever $X$ is unresponsive to $D$, then $X$ will be unresponsive to $D$ in states limited by any $Z$. Also, Properties 4 and 5 imply that limited unresponsiveness is transitive: $X \not\hookleftarrow_Y D$ and $Y \not\hookleftarrow_Z D$ imply $X \not\hookleftarrow_Z D$.

In closing this section, we note that the definition of limited unresponsiveness can be generalized in several ways. In one generalization, we can define what it means for $X \subseteq U$ to be unresponsive to $D$ in states of the world limited by an *instance* of $Y$. Namely, we say that $X$ is unresponsive to $D$ in states limited by $Y = \mathbf{Y}$ if, for all possible states of the world $\mathbf{S}$, and for any two alternatives $\mathbf{D_1}$ and $\mathbf{D_2}$, $Y[\mathbf{S}, \mathbf{D_1}] = Y[\mathbf{S}, \mathbf{D_2}] = \mathbf{Y}$ implies $X[\mathbf{S}, \mathbf{D_1}] = X[\mathbf{S}, \mathbf{D_2}]$. Furthermore, we can imagine generalizations where the possible states of the world are limited by subsets of instances of $Y$, not just a single instance of $Y$.

In a second generalization, we can define what it means for a set of chance variables to be unresponsive to a *subset* of all of the decisions. In particular, given a domain described by $U$ and $D$, we say that $X \subseteq U$ is unresponsive to $D' \subseteq D$ in worlds limited by $Y$ if $X \not\hookleftarrow_{Y \cup (D \setminus D')} D$.

Finally, we can have combinations of these two generalizations. Nonetheless, except for a brief mention of each generalization, we do not pursue them in the remainder of the paper for the sake of simplicity.

## 3 Definition of Cause

Armed with the primitive notion of limited unresponsiveness, we can now formalize our definition of cause.

**Definition 2 (Causes with Respect to Decisions)** *Given a decision problem described by $U$ and $D$, and a variable $x \in U$, the variables $C \subseteq D \cup U \setminus \{x\}$ are said to be causes for $x$ with respect to $D$ if $C$ is a minimal set of variables such that $x \not\hookleftarrow_C D$.*

In our framework, decision variables cannot be caused, because they are under the control of the decision maker. Consequently, we define causes for chance variables only. The definition says that if we can find set of variables $Y$ such that, for all possible states of the world, $x$ can be different for different alternatives only when $Y$ is different, then $Y$ must contain a set of causes for $x$. Our definition of cause departs from traditional usage of the term in that we consider causal relationships relative to a set of decisions. Nonetheless, we find this departure has an important advantage, which we discuss shortly.

As an example of our definition, consider the decision to continue or quit smoking, described by the decision variable $s$ (*smoke*) and the chance variable $l$ (*lung cancer?*). If we believe that $s$ and $l$ are probabilistically dependent, then, by Theorem 1, it must be that $l \hookleftarrow s$. Furthermore, by Property 3, we know that $l \not\hookleftarrow_s s$. Consequently, by Definition 2, we have that $s$ is a cause of $l$ with respect to $s$.

Several consequences of Definition 2 are worth mentioning. First, although cause is irreflexive by definition, it is not always asymmetric. For example, in our story about the coin toss, consider another variable $m$ that represents whether or not the outcome of the coin toss matches our bet $b$. In the story as we have told it, $m$ is a deterministic function of $w$ (*win?*), and vice versa. Consequently, we have $w \not\hookleftarrow_m b$ and $m \not\hookleftarrow_w b$; and so $m$ is a cause of $w$ and $w$ is cause of $m$ with respect to $b$. Note that any hint of uncertainty destroys this symmetry. For example, if there is a possibility that the person tossing the coin will cheat (so that we may lose even if we match), then we can conclude that $m$ is a cause of $w$, but not vice versa.

Second, cause is transitive for single variables. In particular, if $x$ is a cause for $y$ and $y$ is a cause for $z$ with respect to $D$, then $z \hookleftarrow_D$ and (by the transitivity of unresponsiveness) $z \not\hookleftarrow_x D$. Consequently, $x$ is a cause for $z$ with respect to $D$. Note that transitivity does not necessarily hold for causes containing sets of variables, because the minimality condition in Definition 2 may not be satisfied.

Third, $C = \emptyset$ is a set of causes for $x$ with respect to $D$ if and only if $x$ is unresponsive to $D$.

Finally, we have the following theorem, which follows from Definition 2 and several of the properties of limited unresponsiveness given in Section 2.

**Theorem 2** *Given any $x \in U$, if $C$ is a set of causes for $x$ with respect to $D$, and $w \in C \cap U$, then $w$ must be responsive to $D$.*

**Proof:** For any chance variable $w \in C$, let $C' = C \setminus \{w\}$. By the minimality condition in our definition, we have
$$x \hookleftarrow_{C'} D \qquad (1)$$
Suppose that $w \not\hookleftarrow D$. Then, by Property 4, we have
$$w \not\hookleftarrow_{C'} D \qquad (2)$$
Applying Equations 1 and 2 to Property 6, we have that $x \hookleftarrow_C D$, which contradicts that $C$ is a set of causes for $x$ with respect to $D$. $\square$

Let us consider another example of our definition that illustrates an advantage of defining cause with respect to the set of decisions. In the medical-treatment story, we have that $c$ (*cured?*) is responsive to $r$ (*recommendation*), because (among other reasons) in the first row in Table 3, the patient is cured if and only if we recommend the treatment. Furthermore, as we discussed in the previous section, $c$ is unresponsive to $r$ in states limited by $t$ (*taken?*). Consequently, we have that $t$ is a cause of $c$ with respect to $r$.

Now, let us extend this example by imagining that there is some gene that affects how a person reacts to both our recommendation and to therapy. In this situation, it is reasonable for us to assert that the variable $g$ (*genotype?*) is unresponsive to $r$. Thus, by Theorem 2, $g$ cannot be among the causes for any other variable. Someday, however, it may be possible to use retroviral therapy to alter one's genetic makeup. Given an additional decision variable $v$ (*retroviral therapy*), it is reasonable for us to assert that $t$ is responsive to $D = \{r, v\}$ in states limited by $r$, but unresponsive to $D$ in states limited by $\{r, g\}$. In this case, we can conclude that $\{r, g\}$ is a cause for $t$ with respect to $D$. In addition, we can conclude that $\{t, g\}$ is a cause for $c$ with respect to $D$.

Thus, an advantage of defining cause with respect to the set of decisions is that we do not have to attach a causal explanation to dependencies between a variable $x$ and other variables, when we can do nothing to change $x$. In our example, $g$, $t$, and $c$ are probabilistically dependent. Nonetheless, if we cannot do anything to affect genotype, then there is little point in determining whether or not genotype causes treatment and cure; and it is precisely in this case that our definition says it is OK to ignore such questions of cause.

Of course, we sometimes want to be able to assert the existence or nonexistence of causal dependencies outside of a real decision setting. Our definition does not preclude the ability to make such assertions. Namely, there is no reason to require that the decisions $D$ be implementable in practice or at all. If we want to think about whether or not the patient's genotype is a cause for his cure, then we can imagine the retroviral-therapy decision that affects genotype regardless of the availability of the therapy. As another example, if we want to discuss the possibility that gender causes breast cancer, then we can imagine a decision that changes one's gender.

Finally, we can generalize our definition of what it means for a set of *variables* to cause $x$ to a definition of what it means for a set of *instances* to cause $x$. Namely, we say that *instance* **C** *of variables* $C$ *is a cause for* $x \notin C$ *with respect to* $D$ if $C$ is a minimal set of variables such that $x$ is unresponsive to $D$ in states limited by $C = \mathbf{C}$. That is, the instance **C** of $C$ is a cause for $x$ with respect to $D$ if we replace our definition of cause with the weaker requirement that $x$ be unresponsive to $D$ in states limited by $C = \mathbf{C}$. Again, for the sake of simplicity, we do not pursue this generalization in the remainder of the paper.

## 4  Graphical Representation of Cause

Given the known benefits of the Bayesian network for representing conditional independence, we would like a graphical representation of cause and effect. The representation we describe is a special case of an influence diagram. An *influence diagram* for a decision problem described by $U$ and $D$ is a model for that problem having a structural component and a probabilistic component. The *structure* of an influence diagram is a directed acyclic graph containing (square) decision and (oval) chance nodes corresponding to decision and chance variables, respectively, as well as information and relevance arcs. Information arcs, which point to decision nodes, represent what is known at the time decisions are made. Relevance arcs, which point to chance nodes, represent (by their absence) assertions of conditional independence. Namely, for some ordering of the variables, each variable $x$ is probabilistically independent of all preceding variables given the parents of $x$. Associated with each chance node $x$ in an influence diagram are probability distributions that, when combined with the assertions of conditional independence encoded in the structural component, determine the joint probability distribution for $U$ given $D$. A special kind of chance node is the deterministic node (depicted as a double oval). A node $x$ is a *deterministic node* if its corresponding variable is a deterministic function of its parents. Also, an influence diagram may contain a single distinguished node, called a *utility node* that encodes the decision maker's utility for each state of the node's parents. A utility node is a deterministic function of its predecessors and can have no children. Finally, for an influence diagram to be well formed, its decisions must be totally ordered by the influence-diagram structure. (For more details, see Howard [1981].)

In this paper, we concern ourselves neither with the ordering of decision nodes nor the observation of chance variables before making decisions. Therefore, we have no need for information arcs. In addition, although our new concepts apply to models that include a utility node, we do not examine such models, as we can illustrate these concepts with models containing only chance, deterministic, and decision variables. An influence diagram (without information arcs or a utility node) for the medical-treatment problem is shown in Figure 1a.

In Heckerman and Shachter (1994), we showed that an ordinary influence diagram is an inadequate representation of causal dependence. In this section, we discuss a particular kind of an influence diagram, known as an influence diagram in canonical form, that can accurately represent causal relationships.

### 4.1  Mapping Variables and Causal Mechanisms

Before we can describe canonical form, we need to introduce the concept of a *mapping variable.* To understand the concept of a mapping variable, let us reexam-ine Savage's basic formulation of a decision problem. Recall that the chance variables $U$ are a deterministic function of the decision variables $D$ and the state of the world $S$. In effect, each possible state of the world defines a mapping from the decisions $D$ to the chance variables $U$. Thus, $S$ represents all possible mappings from $D$ to $U$. We can characterize $S$ as a mapping variable for $U$ as a function of $D$, and use the suggestive notation $U(D)$ to denote this mapping variable.

Table 4: The mapping variable $t(r)$.

| instance of $t(r)$ | $r$ =take | $r$ =don't take |
|---|---|---|
| 1: complier | $t$ =yes | $t$ =no |
| 2: defier | $t$ =no | $t$ =yes |
| 3: always taker | $t$ =yes | $t$ =yes |
| 4: never taker | $t$ =no | $t$ =no |

In general, given a domain described by $U$, $D$, and $S$, a set of decision variables $Y \subseteq D$, and a set of chance variables $X \subseteq U$, the mapping variable $X(Y)$ is a variable that represents the possible mappings from $Y$ to $X$. Rubin (1978) and Howard (1990) define concepts similar to the mapping variable.

As an example, consider the medical-treatment story. The mapping variable $t(r)$ represents the possible mappings from the decision variable $r$ (*recommendation*) to the chance variable $t$ (*taken?*). In this example, the instances of $t(r)$, shown in Table 4, have a natural interpretation. In particular, the instance where the patient accepts treatment if and only if we recommend it represents a patient who "complies" with our recommendation; the instance where the patient accepts treatment if and only if we recommend against it represents a patient who "defies" our recommendation; and so on.

An important property concerning mapping variables is that, given variables $X$, $Y$, and $X(Y)$, we can always write $X$ as a deterministic function of $Y$ and $X(Y)$. For example, $t$ is a deterministic function of $r$ and $t(r)$; and, more generally, $U$ is a deterministic function of $D$ and $U(D) \equiv S$.

In the discussions that follow, it is important to extend the definition of a mapping variable to include chance variables as arguments. Doing so allows us to decompose the monolithic mapping variable $U(D) \equiv S$ for a domain into a set of variables. For example, consider the medical-treatment story. Given this extension of the mapping-variable definition, we can define the mapping variable $c(t)$ with instances "helped," "hurt," "always cured," and "never cured." Together, the mapping variables $t(r)$ and $c(t)$ describe the possible states of the world $U(D) \equiv S$. (E.g., $t(r) =$"complier" and $c(t) =$"helped" corresponds to state 1 in Table 3.) As we shall see, this decomposition facilitates the graphical representation of causal relationships.

The extension of the mapping-variable definition to include chance variables as arguments is a bit tricky. For example, when the patient is an "always taker", it is impossible to distinguish between the instances "helped" and "always cured" of $c(t)$, because for both recommendations, the patient will accept the treatment. In this sense, the variable $c(t)$ is not well defined.

We can overcome this problem by imagining a decision that allows us to *directly set* $t$ to any of its instances, regardless of the recommendation decision. The key idea in setting this variable *directly* is that we force $t$ to take on a particular instance without changing the instances of any other variables except those that are mandated by the known causal relationships in the domain. For example, assuming the treatment is a drug and that there is no placebo effect, we can directly set $t$ to "taken" by injecting the patient with the drug without his knowledge. In contrast, although we can set $t$ to "taken" by physically forcing the patient to take the drug, this operation may not qualify as a setting of the variable if the patient's conditioned is worsened by the use of force itself.

Pearl and Verma (1991) and Spirtes et al. (1993) discuss the notion of directly setting or manipulating a variable, taking this concept to be primitive. Here, we formally define the notion in terms of limited unresponsiveness.

**Definition 3 (Set Decision)** *Given a domain described by $U$, $D$, and $S$, consider a set of decision variables outside $D$, denoted $\hat{U}$, that contains one decision variable $\hat{x}$ for every $x \in U$, where $\hat{x}$ has alternatives "set $x$ to $k$" for each possible instance $k$ of $x$ and "do nothing." Let $U' = U$, $D' = D \cup \hat{U}$, and $S'$ be an augmentation of the original domain in the sense that, (1) when each $\hat{x} \in \hat{U}$ is set to "do nothing", the realizations in the augmented domain (as a function of $S'$ and $D'$) are the same as those in the original domain, and (2) when $\hat{x} =$ "set $x$ to $k$," then $x$ assumes the state $k$. Then, $\hat{U}$ is said to be a collection of set decision variables for $U$ with respect to $U$, $D$, and $S$ if, for all $Y \subseteq U$ and $Z \subseteq U \cup D$, $x \not\leftarrow_Z D$ in the original domain if and only if $x \not\leftarrow_{Z \cup Y} D \cup \hat{Y}$*[6] *in the augmented domain, where $\hat{Y}$ are the set decisions corresponding to the variables in $Y$.*

For example, in the medical-treatment story, we have that $c \not\leftarrow_t r$. Thus, in the augmented domain, we must have $c \not\leftarrow_t \{r, \hat{t}\}$ for $\hat{t}$ to be a set decision for $t$. It is likely that a decision to secretly inject the patient satisfies this condition (again, provided there is no placebo effect), whereas it is unlikely that a decision to use physical force does. Note that, in general, set decisions need only be hypothesized. They need not be implementable in practice.

---

[6]In writing this expression, we are using the second generalization of the definition of limited unresponsiveness discussed in Section 2. In particular, this expression is equivalent to the statement $x \not\leftarrow_{Z \cup Y \cup (\hat{U} \setminus \hat{Y})} D'$.

**Definition 4 (Setting a Variable)** *Given a decision variable $d$, we* set *that variable by choosing one of its alternatives. Given a chance variable $x$, we* set *that variable by choosing one of the alternatives of $\hat{x}$ other than "do nothing."*

We can now give the general defintion of a mapping variable.

**Definition 5 (Mapping Variable)**
*Given chance variables $X$ and variables $Y$, the* mapping variable $X(Y)$ *is the chance variable that represents all possible mappings from $Y$ to $X$ as we set $Y$ to each of its possible instances.*[7]

There are several important points to be made about mapping variables as we have now defined them. First, as in the more specific case, $X$ is always a deterministic function of $Y$ and $X(Y)$.

Second, additional probability assessments typically are required when introducing a mapping variable into a probabilistic model. For example, two independent assessments are needed to quantify the relationship between $r$ and $t$ in the medical-treatment story; whereas three independent assessments are required for the node $t(r)$. In general, many additional assessments are required. If $X$ has $b$ instances and $Y$ has $a$ instances, then $X(Y)$ has as many as $b^a$ instances. In real-world domains, however, reasonable assertions of independence decrease the number of required assessments. In some cases, no additional assessments are necessary (see, e.g., Heckerman et al. 1994).

Third, although we may not be able to observe a mapping variable directly, we may be able to learn something about it. For example, we can model the decision to continue or quit smoking using the decision variable $s$ (*smoke*), the chance variable $l$ (*lung cancer?*), and the mapping variable $l(s)$. Although we cannot observe $l(s)$, we can imagine a test that measures the susceptibility of someone's lung tissue to lung cancer in the presence of tobacco smoke. Given the result of such a test, we can update our probability distribution over $l(s)$.

Fourth, we have the following theorem and corollaries.

**Theorem 3 (Mapping Variable)** *Given a decision problem described by $U$ and $D$, variables $X \subseteq U$, and variable sets $W$, $Y$, and $Z$ that are all subsets of $U \cup D$, $X(W) \not\leftarrow_{Z \cup Y} D$ if and only if $X(W \cup Y) \not\leftarrow_Z D$.*

**Proof:** $X(W \cup Y)$ represents all possible mappings from $W \cup Y$ to $X$. By the definition of a mapping variable, $X(W \cup Y) \not\leftarrow_Z D$ if and only if $X(W) \not\leftarrow_{Z \cup Y} D \cup \hat{Y}_U$, where $Y_U = Y \cap U$ is the set of chance variables in $Y$. Likewise, by the definition of a set decision, $X(W) \not\leftarrow_{Z \cup Y} D \cup \hat{Y}_U$ if and only if $X(W) \not\leftarrow_{Z \cup Y} D$. □

**Corollary 4 (Mapping Variable)** *Given a decision problem described by $U$ and $D$, variables $X \subseteq U$, and $Y \subseteq U \cup D$, $X \not\leftarrow_Y D$ if and only if $X(Y) \not\leftarrow D$.*

For example, in the medical treatment story, we have $c \not\leftarrow_t r$ and $c(t) \not\leftarrow r$. Roughly speaking, Corollary 4 says that $X$ is unresponsive to $D$ in states limited by $Y$ if and only if the way $X$ depends on $Y$ does not depend on $D$. This equivalence provides us with an alternative set of conditions for cause.

**Corollary 5 (Cause)** *Given a decision problem described by $U$ and $D$, and a chance variable $x \in U$, the variables $C \subseteq D \cup U \setminus \{x\}$ are* causes *for $x$ with respect to $D$ if $C$ is a minimal set of variables such that $x(C) \not\leftarrow D$.*

We can think of $x(C)$—where $C$ are causes for $x$—as a *causal mechanism* that relates $C$ and $x$. For example, suppose chance variables $i$ and $o$ represent the voltage input and output, respectively, of an inverter in a logic circuit. Given a decision $d$ to which $i$ responds, we can assert that $\{i\}$ is a cause for $o$. In this example, the mapping variable $o(i)$, represents the mapping from the inverter's inputs to its outputs. That is, this mapping variable represents the state of the inverter itself.

**Definition 6 (Causal Mechanism)** *Given a decision problem described by $U$ and $D$ and a chance variable $x \in U$ that is responsive to $D$, a* causal mechanism *for $x$ with respect to $D$ is a mapping variable $x(C)$ where $C$ are causes for $x$ with respect to $D$.*

Thus, we have the following consequence of Corollary 4.

**Corollary 6 (Causal Mechanism)** *If $x(C)$ is a causal mechanism for $x$ with respect to $D$, then $x(C)$ is unresponsive to $D$.*

### 4.2 Canonical Form Influence Diagrams

We can now define what it means for an influence diagram to be in canonical form.

**Definition 7 (Canonical Form)** *An influence diagram for a decision problem described by $U$ and $D$ is said to be in* canonical form *if (1) all chance nodes that* are responsive to $D$ are descendants of one or more decision nodes and (2) all chance nodes that are descendants of one or more decision nodes are deterministic nodes.

---
[7] There are some technical details involved with the definition of a mapping variable when particular instances of $Y$ are not possible or not possible for particular instances of $D$. Although all of the results given here are true in general, we omit these special cases for simplicity in presentation.

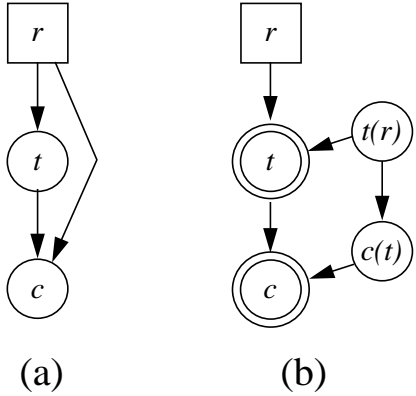

Figure 1: (a) An influence diagram for the medical-treatment story. (b) A corresponding influence diagram in canonical form.

An immediate consequence of this definition is that any chance node that is not a descendant of decision node must be unresponsive to $D$.

We can construct an influence diagram in canonical form for a given problem by including in the influence diagram a causal mechanism for every variable that is responsive to the decisions. In doing so, we can make every responsive variable a deterministic function of a set of its causes and the unresponsive causal mechanism. For example, consider the medical-treatment story as depicted in the influence diagram of Figure 1a. The variables $t$ and $c$ are responsive to $r$, but their corresponding nodes are not deterministic. Consequently, this influence diagram is not in canonical form. To construct a canonical form influence diagram, we introduce the mapping variables $t(r)$ and $c(t)$, as shown in Figure 1b. The responsive variables are now deterministic; and the mapping variables are unresponsive to the decision. This example illustrates an important point: Causal mechanisms may be probabilistically dependent. We return to this issue in Section 4.3.

In general, we can construct an influence diagram in canonical form for the decision problem $U$ and $D$ as follows.

**Algorithm 1 (Canonical Form)**

1. Add a node to the diagram corresponding to each variable in $U \cup D$

2. Order the variables $x_1, \ldots, x_n$ in $U$ so that the variables unresponsive to $D$ come first

3. For each variable $x_i \in U$ that is unresponsive to $D$,

    (a) Add a causal-mechanism chance node $x_i(C_i)$ to the diagram,
    where $C_i \subseteq D \cup \{x_1, \ldots, x_{i-1}\}$

    (b) Make $x_i$ a deterministic node with parents $C_i$ and $x_i(C_i)$

4. Assess dependencies among the variables that are unresponsive to $D$

This algorithm is well defined. In particular, it is always possible to find a $C_i$ satisfying the condition in step 3a, because $x_i \not\leftarrow_D D$ by Property 3.

Also, the structure of the of the constructed influence diagram is valid. Namely, by Corollary 6, all causal mechanisms added in step 3 are unresponsive to $D$. Thus, suppose we identify the relevance arcs and deterministic nodes by using a variable ordering where the nodes in $D$ are followed by the unresponsive nodes (including the causal mechanisms), which are in turn followed by the responsive nodes in the order specified at step 2. Then, (1) we would add no arcs from $D$ to the unresponsive nodes by Theorem 1 (and the algorithm adds none); (2) we would add arcs among the unresponsive nodes as described in step 4; and (3) for every responsive variable $x_i$, we would make $x_i$ a deterministic node (as described in step 3b) by definition of a mapping variable.

Furthermore, the structure that results from Algorithm 1 will be in canonical form. In particular, because there are no arcs from $D$ to the unresponsive nodes, only responsive variables can be descendants of $D$. In addition, by Theorem 2, we know that every responsive node is a descendant of $D$, and (by construction) a deterministic node.

To illustrate the algorithm, consider the medical-treatment story as depicted by the influence diagram in Figure 2a where the variable $g$ (genotype?) is represented explicitly. To construct an influence diagram in canonical form for this problem, we first add the variables $\{r, g, t, c\}$ to the diagram and choose the ordering $(g, t, c)$. Both $t$ and $c$ are responsive to $D = \{r\}$, and have causes $r$ and $t$, respectively. Consequently, we add causal mechanisms $t(r)$ and $c(t)$ to the new diagram, and make $t$ a deterministic function of $r$ and $t(r)$ and $c$ a deterministic function of $t$ and $c(t)$. Finally, we assess the dependencies among the unresponsive variables $\{g, t(r), c(t)\}$, adding arcs from $g$ to $t(r)$ and $c(t)$ under the assumption that the causal mechanisms are conditionally independent given $g$. The resulting canonical form influence diagram is shown in Figure 2b.

From our construction, it follows that every responsive variable $x_i$ has at least one set of causes explicitly encoded in the diagram ($C_i$). That is, a canonical form influence diagram constructed as in Algorithm 1 accurately represents a set of causes for every variable having a nonempty set of causes. In this sense, we find canonical form to be an adequate representation of cause.

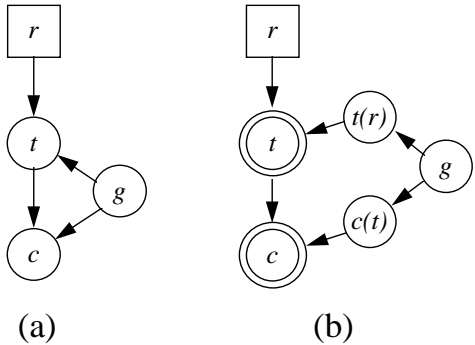

Figure 2: (a) Another influence diagram for the medical-treatment story. (b) A corresponding influence diagram in canonical form.

Canonical form is a generalization of Howard Canonical Form, which was developed by Howard (1990) to facilitate the computation of value of information.

### 4.3 Pearl's Causal Theory

There is a close relationship between the canonical form influence diagram and Pearl's causal theory (Pearl and Verma, 1991; Pearl, 1995). In fact, as we now demonstrate, a causal theory is a special case of canonical form.[8]

Pearl takes causation to be a primitive notion, and defines a *causal model for variables* $U$ to be a directed acyclic graph where each node corresponds to a variable in $U$ and each nonroot node is caused by its parents. Each variable in his analysis plays a dual role of chance and decision variable. In particular, a variable may be observed or directly set to a particular instance. As mentioned, Pearl takes the concept of directly setting a variable to be a primitive.

Given a causal model for $U$, Pearl goes on to define a causal theory for $U$. Here, we express his definition in the language of influence diagrams. Let $\mathcal{M}(U)$ be a causal model for $U$. Let $Pa(x)$ denote the parents of $x$ in $\mathcal{M}(U)$, which by definition are causes for $x$. A *causal theory for* $U$ *based on* $\mathcal{M}(U)$, which we denote $\mathcal{T}(U)$, is an influence diagram described as follows. For each variable $x_i \in U$, $i = 1, \ldots, n$, $\mathcal{T}(U)$ contains a corresponding chance variable $x_i$, a set decision $\hat{x}_i$ for

$x_i$, and a chance variable $\epsilon_i$, which Pearl calls a *disturbance variable*. Furthermore, in the influence diagram $\mathcal{T}(U)$, only the chance nodes $x_i$ have parents. In particular, each $x_i$ is a deterministic function of $Pa(x_i)$, $\hat{x}_i$, and $\epsilon_i$, where (1) if $\hat{x}_i = k$ then $x_i = k$, and (2) if $\hat{x}_i =$ "do nothing" then $x = f_i[Pa(x_i), \epsilon_i]$ for some deterministic function $f_i$. Note that, in a causal theory, disturbance variables are mutually independent by definition.

Now, in our framework, suppose we have a set of chance variables $U$ and a corresponding collection of set decisions $\hat{U}$ for $U$ with respect to $U$. In addition, suppose that, for all $x_i$, $Pa(x_i) \cup \{\hat{x}_i\}$ is a set of causes for $x_i$ with respect to $\hat{U}$. When we construct an influence diagram in canonical form as described in Algorithm 1 using an ordering consistent with the causal model $\mathcal{M}(U)$, we can obtain an influence diagram where each variable $x_i$ is a deterministic function of $\hat{x}_i$, $Pa(x_i)$, and the causal mechanism $x_i(Pa(x_i), \hat{x}_i)$. Given the definition of a set decision, we can simplify each such relationship by writing $x_i$ as a deterministic function of $\hat{x}_i$, $Pa(x_i)$, and the variable $x_i(Pa(x_i))$, where $x_i(Pa(x_i))$ represents the possible mappings from $Pa(x_i)$ to $x_i$ when $\hat{x}_i$ is set to "do nothing." If we identify each mapping variable $x_i(Pa(x_i))$ with Pearl's disturbance variable $\epsilon_i$, then we obtain an influence diagram identical to the causal theory $\mathcal{T}(U)$, with the exception that the mapping variables in this influence diagram may be dependent.[9]

The fact that disturbance variables must be independent in a causal theory does not necessarily limit the expressiveness of a causal theory. Such dependencies often disappear when hidden common causes are introduced. Furthermore, the assumption that causal mechanisms are independent has the convenient consequence that the a causal model for $U$ can be interpreted as a Bayesian network in the traditional sense (Spirtes et al., 1993; Pearl, 1995). That is, if variables $X$ and $Y$ are d-separated by $Z$ in the causal model, then $X$ and $Y$ are conditionally independent given $Z$ according to the causal theory.

Nonetheless, the fact that we can use canonical form to represent causes locally—that is, we represent causes only when they are relevant to the decisions at hand—makes canonical form a more efficient representation than the causal theory. For example, to represent the relationships in Figure 2b using a causal theory, we would introduce causal-mechanism variables $t(r, g)$ and $c(t, g)$. Assuming $r, g, t$ and $c$ are binary variables, both mapping variables in the causal theory would have 16 instances. In contrast, both mapping variables in Figure 2b have only four instances. Consequently, the nodes $g$, $t(r,g)$ and $c(t,g)$ in the causal-theory representation require 31 probabilities in total, whereas the nodes $g$, $c(s)$, and $v(d)$ in the canonical-form representation require only 13 probabilities in total.

---

[8] We note that Pearl's causal theory and the *pseudo-indeterministic system* of Spirtes et al. (1993) are very similar, and many of the remarks in this section apply to the latter representation as well.

[9] At first glance, there appears to be another difference between the two representations. Namely, the disturbance variables in $\mathcal{T}(U)$ are assumed to be merely independent of the set decisions $\hat{U}$; whereas, in the canonical form influence diagram, the mapping variables are also unresponsive to the set decisions. A careful reading of Pearl's work, however, suggests that the disturbance variables must in fact be unresponsive to the set decisions.

## 5 Conclusions and Future Work

We have presented a precise definition of cause and effect in terms of the more fundamental notion of unresponsiveness. Our definition departs from the traditional view of causation in that our causal assertions are made relative to a set of decisions. As a consequence, our definition allows for models where only some dependencies have a causal explanation. We have shown how these properties can make the representation and manipulation of causal relationships more efficient.

In addition, we have examined the graphical encoding of causation. We have shown how the ordinary influence diagram is inadequate as a graphical representation of cause, but that the canonical form influence diagram is always an accurate language for causal dependence. Also, we have described the relationship between Pearl's causal theory and canonical form influence diagrams.

An important aspect of causality that we have barely touched upon in this paper is the notion of time. For example, what if our alternatives consists of *reactive plans*, where observations are interspersed with actions? More generally, what happens when system variables change in time? We will explore these issues and others in a sequel to this paper.

## Acknowledgments

We thank Jack Breese, Tom Chavez, Max Chickering, Eric Horvitz, Ron Howard, Christopher Meek, Judea Pearl, Mark Peot, Glenn Shafer, Peter Spirtes, Patrick Suppes, and anonymous reviewers for useful comments.